\documentclass[runningheads]{llncs}
\usepackage[T1]{fontenc}
\usepackage{graphicx} 
\usepackage{epsfig} 
\usepackage{mathptmx} 
\usepackage{times} 
\usepackage{amsmath} 
\usepackage{amssymb}  
\usepackage[mathscr]{eucal}
\usepackage[noend]{algpseudocode}
\usepackage{subfigure}
\usepackage{algorithmicx,algorithm}
\usepackage{multirow} 
\usepackage{booktabs} 
\usepackage{array}
\usepackage{hyperref}
\usepackage{marvosym}
\usepackage{orcidlink}
\usepackage{textcomp}
\hypersetup{hypertex=true,
	colorlinks=true,
	linkcolor=blue,
	anchorcolor=blue,
	citecolor=blue}
\usepackage{graphicx}

\begin{document}
\title{StrokeNet: Unveiling How to Learn Fine-Grained Interactions in Online Handwritten Stroke Classification}
\titlerunning{StrokeNet}

\author{Yiheng Huang\textsuperscript{1*\textdagger}\textsuperscript{\orcidlink{0009-0007-9527-6569}} \and
Shuang She\textsuperscript{2*} \and
Zewei Wei\textsuperscript{2} \and
Jianmin Lin\textsuperscript{2}\textsuperscript{(\Letter)} \and 
Ming Yang\textsuperscript{2} \and
Wenyin Liu\textsuperscript{1}\textsuperscript{(\Letter)}\textsuperscript{\orcidlink{0000-0002-6237-6607}}
}
\institute{College of Computer Science and Technology, Guangdong University of Technology, Guangzhou 510006, People’s Republic of China \email{huangyiheng.gdut@gmail.com,liuwy@gdut.edu.cn} \and
CVTE Research
\\
\email{\{sheshuang,weizewei,linjianmin,yangming\}@cvte.com}}

\renewcommand{\thefootnote}{\fnsymbol{footnote}}
\footnotetext[1]{These authors contributed equally to this work.}
\footnotetext[4]{This work is done during Yiheng Huang’s internship at CVTE Research.}

\maketitle              
%
\begin{abstract}
Stroke classification remains challenging due to variations in writing style, ambiguous content, and dynamic writing positions. 
The core challenge in stroke classification is modeling the semantic relationships between strokes. Our observations indicate that stroke interactions are typically localized, making it difficult for existing deep learning methods to capture such fine-grained relationships. Although viewing strokes from a point-level perspective can address this issue, it introduces redundancy. However, by selecting reference points and using their sequential order to represent strokes in a fine-grained manner, this problem can be effectively solved.
This insight inspired StrokeNet, a novel network architecture encoding strokes as reference pair representations (points + feature vectors), where reference points enable spatial queries and features mediate interaction modeling. Specifically, we dynamically select reference points for each stroke and sequence them, employing an Inline Sequence Attention (ISA) module to construct contextual features. To capture spatial feature interactions, we devised a Cross-Ellipse Query (CEQ) mechanism that clusters reference points and extracts features across varying spatial scales. Finally, a joint optimization framework simultaneously predicts stroke categories via reference points regression and adjacent stroke semantic transition modeling through an Auxiliary Branch (Aux-Branch).
Experimental results show that our method achieves state-of-the-art performance on multiple public online handwritten datasets. Notably, on the CASIA-onDo dataset, the accuracy improves from 93.81$\%$ to 95.54$\%$, demonstrating the effectiveness and robustness of our approach.

\vspace{-0.55em}
\end{abstract}

\vspace{-1.85em}

\section{Introduction}

The ubiquitous digitization of documentation has amplified the demand for reliable online handwritten stroke classification systems, particularly as some challenges remain: variations in writing styles, speed, and contextual factors \cite{artz2020taking,polotskyi2020spatio}.
Effective stroke classification plays a crucial role in various domains \cite{indermuhle2010iamondo,awal2011first,bresler2014recognition}. For example \cite{yang2021casia}, in online education, it enables automated grading and personalized feedback, improving the assessment of handwritten assignments and enhancing the learning experience.

\begin{figure}[t]
\centering
\subfigure[Stroke level visualization]{
    \label{fig1(a)}    
    \includegraphics[height=5.8cm,width=5.8cm]{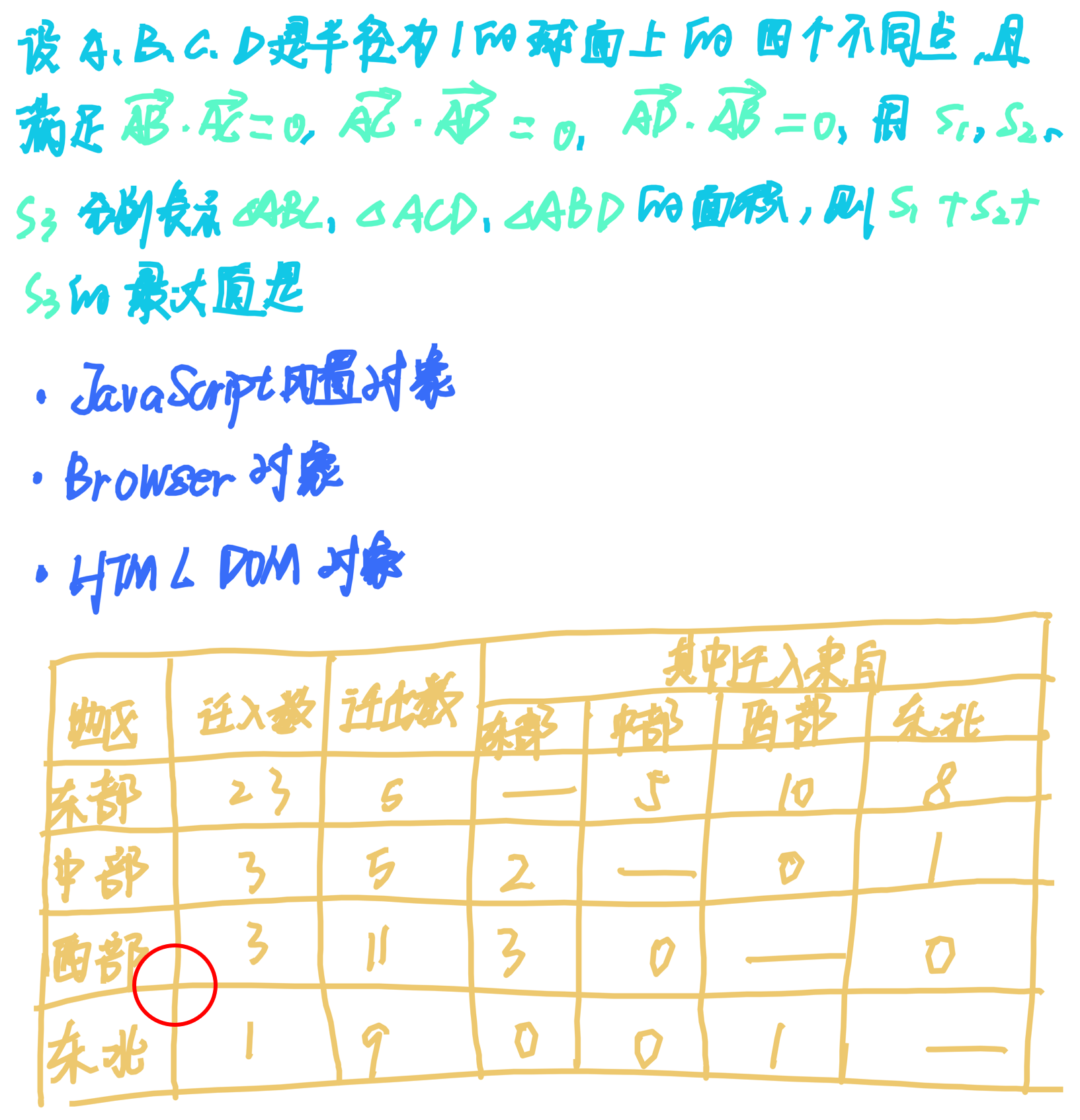}   
}
\subfigure[Point level visualization]{
    \label{fig1(b)}    
    \includegraphics[height=5.8cm,width=5.8cm]{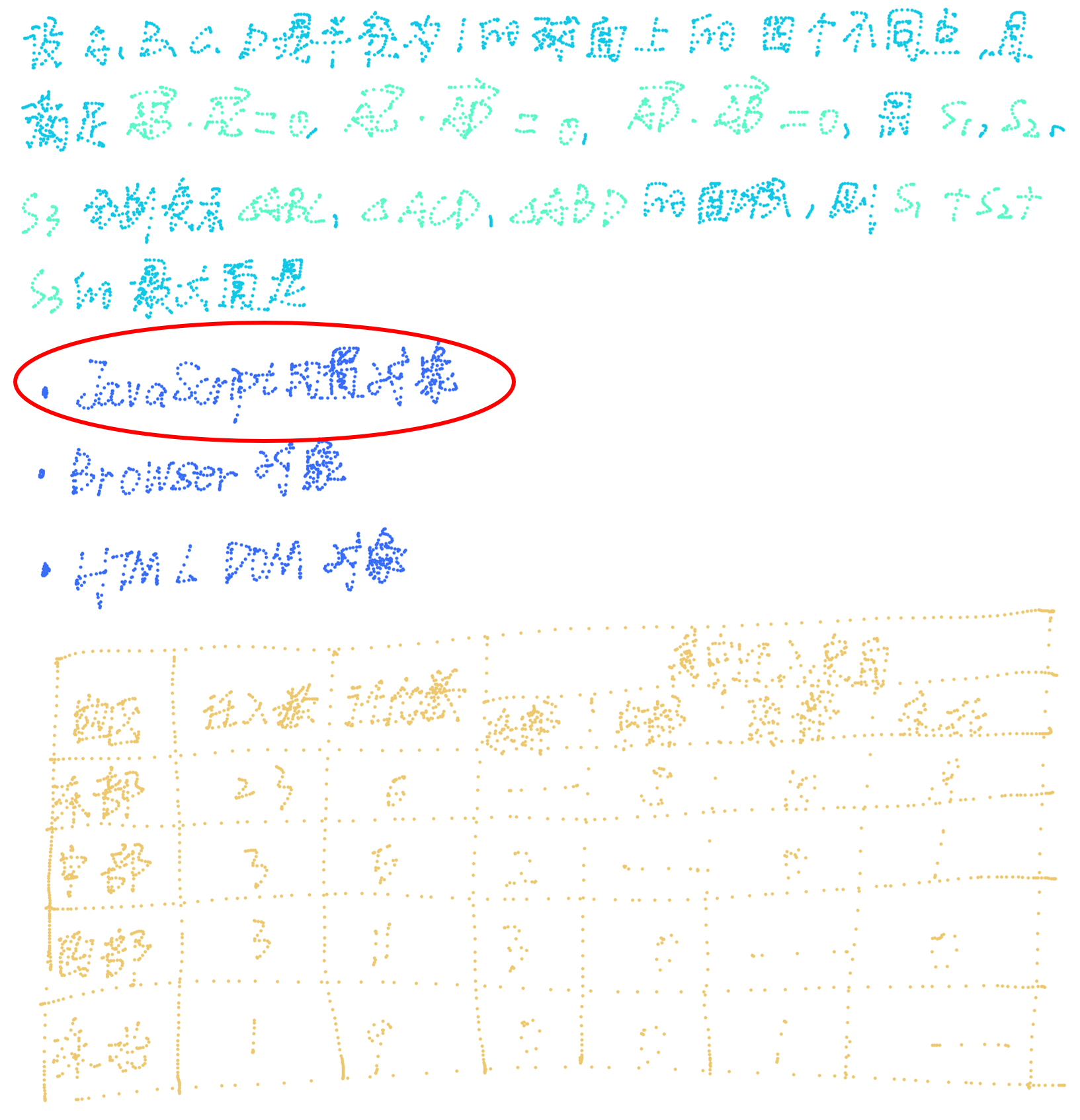}   
}
\vspace{-0.75em}
\caption{ Comparison of stroke and point levels to demonstrate the locality of stroke interactions and the redundancy of points within strokes.}
\label{fig1}
\vspace{-2.0em}
\end{figure}
With the advancement of machine learning, various methods have been proposed for online handwritten stroke classification, including Probabilistic Graphical Model (PGM), Recurrent Neural Network (RNN), Graph Neural Network (GNN), and Transformers. PGMs \cite{bishop2004distinguishing,zhou2007text,ye2016joint} provide interpretable probabilistic frameworks but suffer from limited expressiveness under complex stroke variations due to handcrafted constraints \cite{delaye2014contextual}. RNNs \cite{indermuhle2012mode,van2016combination} excel at temporal sequence modeling, but fail to capture crucial spatial relationships, limiting their ability to represent stroke structures accurately \cite{grygoriev2021hcrnn}. GNNs \cite{ye2019contextual,ye2020contextual} encode spatial structures by representing strokes as graphs and exploit relational information. However, their performance depends heavily on the quality of graph construction, which may introduce biases and affect robustness \cite{yao2024long}. Transformers \cite{mustafid2023iamonsense,zheng2023sketch} capture global dependencies through Self-Attention but require predefined positional encoding and are computationally intensive \cite{liu2024transformer}. These limitations highlight the need for models that effectively integrate both spatial and temporal information while being computationally efficient.

In our study of online handwritten stroke classification, we observed pronounced spatial locality in stroke interactions, as highlighted by the red circle in Fig.~\ref{fig1(a)}. For example, interactions between table lines and elements often concentrate in this area. This observation suggests that stroke-node-based relational representations are insufficient for capturing fine-grained local interactions. Therefore, it is necessary to explore higher-granularity stroke representations. Further analysis of point-level visualization results, as shown by the red circle in Fig.~\ref{fig1(b)}, reveals that directly using all discrete points per stroke increases granularity but also leads to redundancy and significantly higher computational complexity. Consequently, to achieve effective and non-redundant stroke representation, it is crucial to explore a more representative selection strategy.

Inspired by point cloud methods \cite{qi2017pointnet,qi2017pointnet++}, we propose a novel framework called StrokeNet that encodes strokes into a reference pair representations (points + feature vectors). Our approach begins by selecting centroids from these reference points using the Farthest Point Sampling (FPS) algorithm. Subsequently, Cross-Ellipse Query (CEQ) are performed along both X and Y axes to cluster reference points around each centroid, facilitating local interactions among features within these clusters. Through multiple levels feature abstraction, we progressively expand the interaction scope to achieve a comprehensive representation of all strokes. To maximize spatial information utilization, we employ regression techniques on reference points to reconstruct complete stroke representations, thereby enhancing classification accuracy.

To reduce redundancy and improve representativeness, we introduce a dynamic point selection strategy that effectively selects reference point sequences based on cumulative length and thresholds. Additionally, we design an Inline Sequence Attention (ISA) module to efficiently construct reference features. This module calculates the inline relationship between the reference point sequence and the stroke sequence, encoding stroke context information into the feature space. By employing a joint optimization framework that simultaneously predicts stroke categories and adjacent stroke semantic transformations in the Auxiliary Branch (Aux-Branch), we can further reduce bias caused by distant reference points within the grouping area. Extensive experiments on multiple public online handwritten datasets demonstrate our approach's effectiveness and robustness, outperforming existing methods.

In summary, our main contributions are as follows:  
\begin{itemize}
    \item  To our knowledge, this is the first time that point cloud solutions have been applied to online handwritten stroke classification tasks, and has pioneered a new paradigm for solving them.

     \item  Propose a dynamic reference point selection strategy and design an Inline Sequence Attention (ISA) module to construct reference point-feature pairs. In addition, a hierarchical grouping scheme utilizes spatial retrieval relationships to enhance feature interaction, expand the receptive field, and improve stroke-level classification accuracy.

    \item  Introduce a Cross-Ellipse Query (CEQ) to effectively capture interactions in elements with unbalanced aspect ratios. Furthermore, an Auxiliary Branch (Aux-Branch) is designed to supervise semantic transitions between consecutive strokes, ensuring consistency in stroke semantics.
    
    \item  Extensive experiments on multiple public datasets demonstrate the superiority of our method in accuracy and generalization. Notably, on the CASIA-onDo dataset featuring the most complex layout elements, our approach improves accuracy from 93.81$\%$ to 95.54$\%$, validating its effectiveness and robustness.
\end{itemize}

\section{Related Work}
\subsection{Online Handwritten Stroke Classification}
Stroke classification in online documents aims to predict the category of each stroke by utilizing information such as sequential stroke points, writing order, and layout position collected from the screen. Based on stroke association information, existing methods can be categorized into two types: isolated classification and contextual classification. Isolated classification focuses on extracting geometric features of strokes and classifying each one independently into semantic categories. For example, Namboodiri et al. \cite{namboodiri2004robust} used approximation functions for stroke representation, while Emanuel et al. \cite{indermuhle2010iamondo} used an SVM classifier to separate strokes based on features like length and curvature. Although these isolated methods are computationally efficient, they fundamentally limit classification accuracy due to the lack of contextual feature integration.

Bishop et al. \cite{bishop2004distinguishing} and Delaye et al. \cite{delaye2014contextual} explored temporal and contextual relationships  in stroke sequences using Hidden Markov Model (HMM) and Conditional Random Field (CRF). These Probabilistic Graphical Model (PGM) methods improve stroke classification by incorporating temporal dependencies and contextual modeling, but they still encounter challenges related to accuracy and computational cost. On the other hand, Recurrent Neural Network (RNN) have been utilized in several studies \cite{indermuhle2012mode,van2016combination} to capture both global and local contexts within stroke sequences for classification. Grygoriev et al. \cite{grygoriev2021hcrnn} proposed a hierarchical LSTM architecture to model the hierarchical structure of handwritten documents. However, due to the sequential nature of RNN, these models struggle to effectively integrate spatial information, resulting in incomplete semantic understanding and classification ambiguities. 

Given the suitability of graph structures for structured data, Graph Neural Network (GNN) have been considered for stroke classification. Ye et al. \cite{ye2019contextual} modeled documents as graphs and used an attention mechanism to update nodes, but ignored edge features in obtaining attention weights. Ye et al. \cite{ye2020contextual} improved the edge update process using edge pooling techniques, but the receptive field remains limited and suffers from redundancy among adjacent node information. Yao et al. \cite{yao2024long} proposed node and edge clustering graph pooling and un-pooling operations, expanding the model’s receptive field to capture long-term dependencies. However, the cumbersome graph construction and the corresponding hand-crafted features limit the generalization of GNN.

Additionally, some works \cite{mustafid2023iamonsense,zheng2023sketch} have attempted to introduce the Transformer architecture into trajectory studies. Liu et al. \cite{liu2024transformer} designed a hierarchical encoding scheme with a polar coordinate system for the standard Transformer model to precisely model the spatial relationships of strokes. While achieving good accuracy performance, it relies heavily on specific encodings and requires significant computational resources.

In this paper, we abandon complex graph construction frameworks and handcrafted features, proposing a simple yet effective network for stroke classification. We do not deliberately steer away from the trends of online stroke classification research, but strive for an inherently simple and more intuitive StrokeNet analysis architecture.

\vspace{-0.7em}
\subsection{ Point Cloud Analysis}
Point cloud data processing primarily relies on two approaches. Given that point cloud data is inherently irregular and unordered, some studies have converted it into an ordered grid network \cite{simonyan2014very,zhao2023divide,sun2023superpoint,jiang2020pointgroup}, followed by applying 3D sparse convolution. This method enhances understanding and leverages efficient processing capabilities of 3D images and voxel data, but introduces information loss, reducing detail accuracy \cite{yang2019std}. 

To address this, researchers have proposed direct processing methods for raw point clouds. Qi et al. \cite{qi2017pointnet} introduced PointNet, a framework that processes point sets by learning spatial encodings for each point and aggregating individual features into global features. Building on this, Qi et al. \cite{qi2017pointnet++} developed PointNet++, which uses a hierarchical neural network to recursively apply point networks to nested partitions, capturing local geometric structures. Due to its effectiveness in local representation and multi-scale information extraction, PointNet++ has become a cornerstone of modern point cloud processing \cite{chen2021pointnet++,ma2022rethinking,qian2022pointnext}. 

Our approach follows the design principles of PointNet++, while also incorporating the inherent prior knowledge of strokes. Unlike traditional point clouds that are disordered, the points within handwritten strokes inherently exhibit aggregative properties. More importantly, handwritten data contains a unique sequential order. Given this sequential nature, connecting only a few reference points is sufficient to effectively represent a stroke. Consequently, our goal is to develop a network architecture that is better suited for ordered stroke writing tasks.

\begin{figure}[t]
\includegraphics[width=\textwidth]{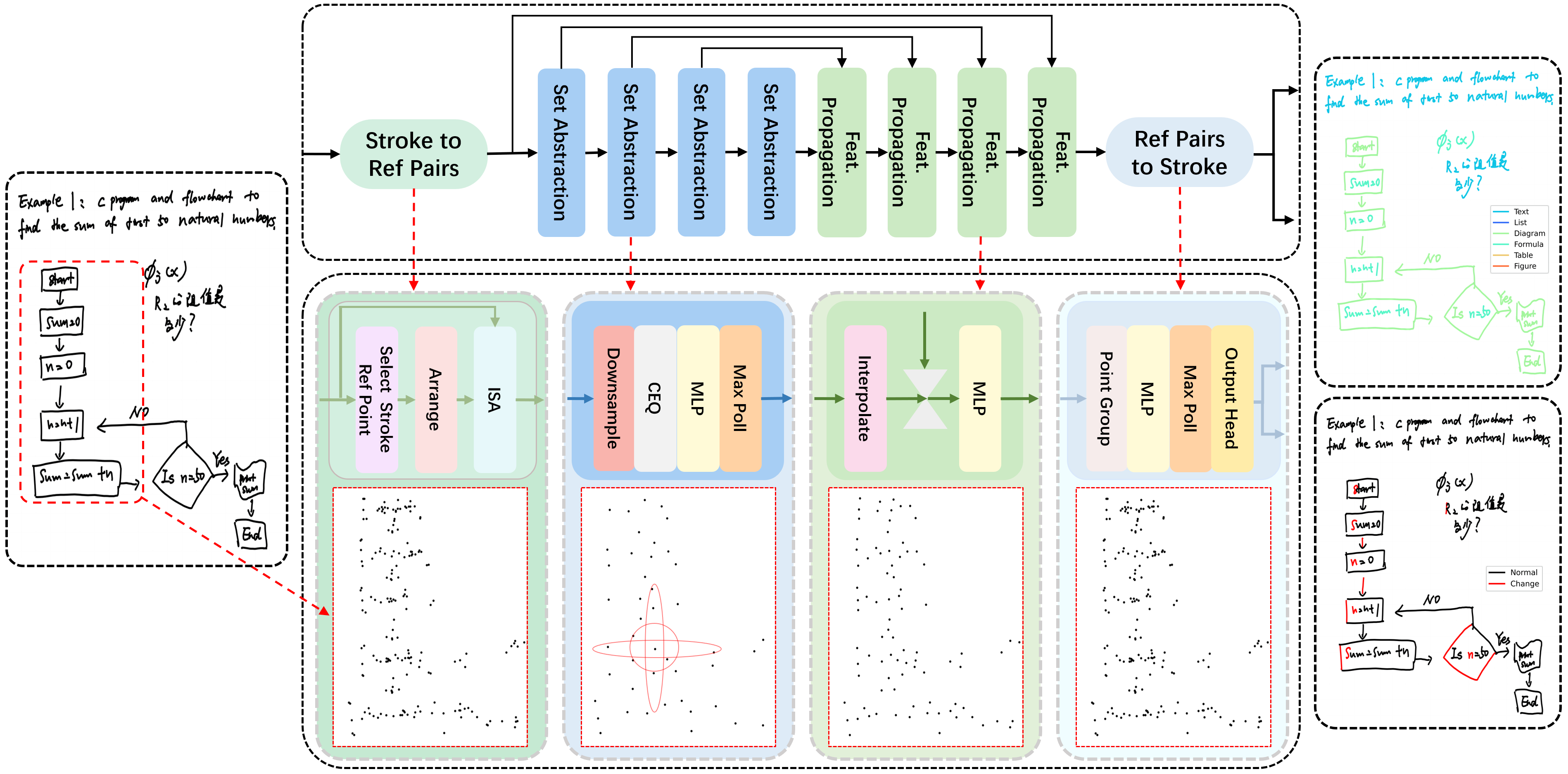} 
\caption{Overview of StrokeNet.} \label{fig2}
\vspace{-1.8em}
\end{figure}

\begin{figure}[t]
\centering
\subfigure[Inline Sequence Attention]{
    \label{fig_3(a)}
    \includegraphics[height=6.6cm,width=5.62cm]{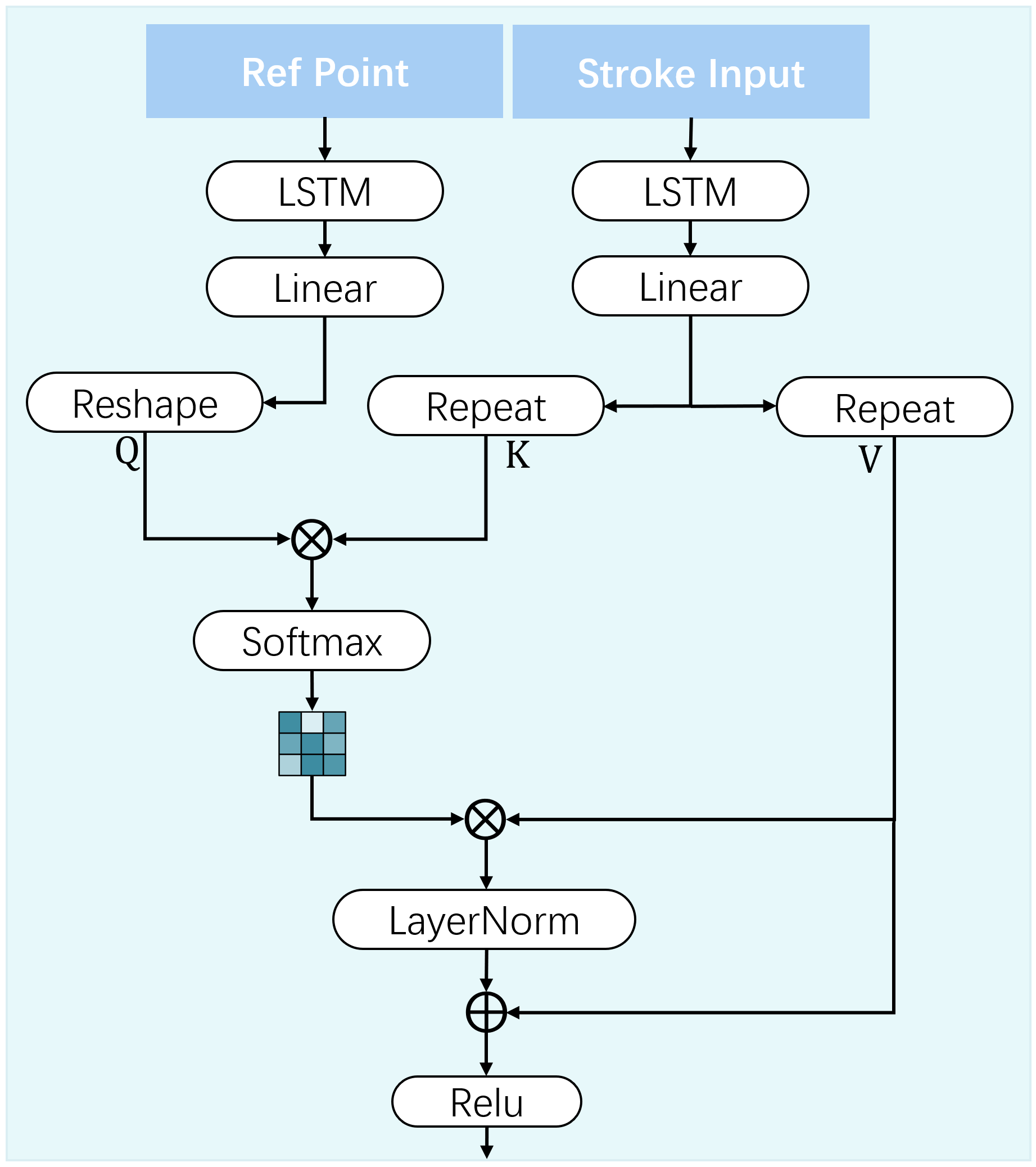}}
\subfigure[Cross-Ellipse Query]{
    \label{fig_3(b)}
    \includegraphics[height=6.6cm,width=6.25cm]{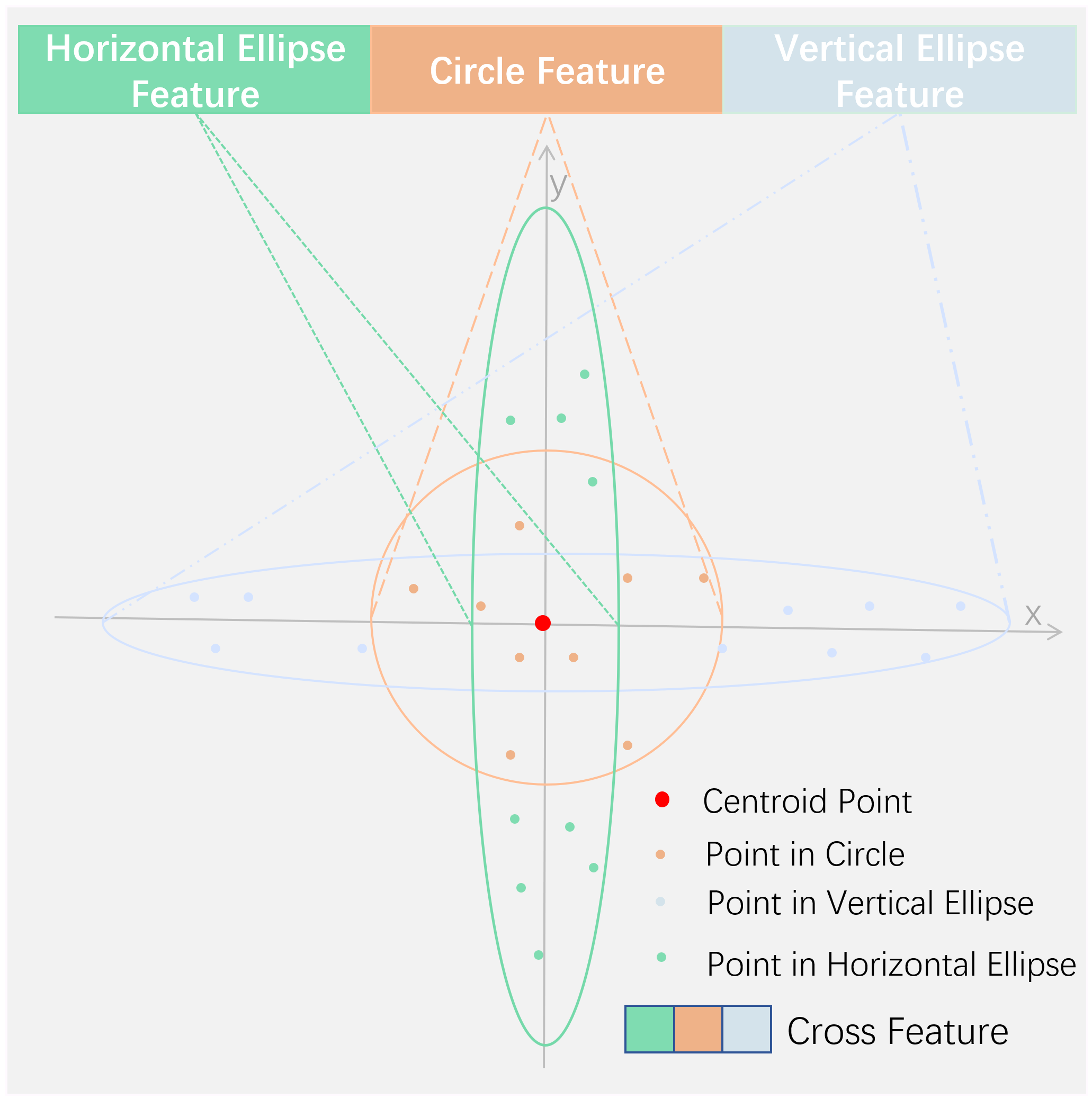}
}
\vspace{-1.2em}
\caption{ Structures of the Inline Sequence Attention (ISA) and Cross-Ellipse Query (CEQ). }
\label{fig3}
\vspace{-1.6em}
\end{figure}

\section{Method}
\textbf{Problem Formulation.}
We are given a set of labeled online handwritten documents. Each document consists of a sequence of strokes $\{S_i,i=1,\ldots,M\}$, where $M$ represents the number of strokes. Each stroke $S_i$ is composed of ordered trajectory points $\{P_i, i = 1,\ldots,k\}$ represented by $(x, y)$-coordinates, where $k$ represents the number of points, and each stroke has a single semantic label $Y_i \in \{1,\ldots,T\}$, where $T$ denotes the number of predefined categories. Our goal is to develop a model based on these training documents that can predict the labels of strokes in testing documents with high accuracy. In our StrokeNet, each stroke adaptively selects different numbers of reference points $R_P$ and extracts corresponding features $R_F$ at the same time, which are jointly constructed as the input reference pair $(R_P, R_F)$ of the model. Thus, online handwritten stroke classification can be represented as a mapping function $\psi$, which assigns each pair in $R$ to an $T$-dimensional semantic label space:
{
	\setlength\abovedisplayskip{0.2cm}
	\setlength\belowdisplayskip{0.2cm}
	\begin{gather}
        \psi : (R_P,R_F) \longmapsto \mathbf{Y}^T 
	\end{gather}
}

\textbf{Process Introduction.} 
As shown in Fig.~\ref{fig2}, we propose StrokeNet, which dynamically selects reference points based on predefined rules and employs an attention-based mechanism to construct reference features. The network follows an encoder-decoder architecture: the encoder aggregates local features using Set Abstraction, the decoder restores reference points through interpolation and skip connections and finally aggregates reference points within each stroke to generate the classification output.

\subsection{Stroke to Reference Pairs Input}
\textbf{Select Stroke Reference Point.} 
To efficiently extract the reference points $R_P$ from each stroke $S_i$, we propose a dynamic selection algorithm. This algorithm utilizes predefined thresholds and minimum reference point counts to select representative points based on segment lengths.

First, we compute the distances $d_i$ between successive points and the cumulative lengths $L_i$. The distance between adjacent points is defined as:
{
	\setlength\abovedisplayskip{0.2cm}
	\setlength\belowdisplayskip{0.2cm}
	\begin{equation}
		\begin{split}
            d_i = \sqrt{ (x_{i+1} - x_i)^2 + (y_{i+1} - y_i)^2 }
		\end{split}	
	\end{equation}
}The cumulative length up to the $i$-th point is defined as:
{
	\setlength\abovedisplayskip{0.2cm}
	\setlength\belowdisplayskip{0.2cm}
	\begin{equation}
		\begin{split}
            L_i = \sum_{j=0}^{i} d_j, \quad L_1 = 0
        \end{split}	
	\end{equation}
}To dynamically select reference points, we initially designate the first point as a reference point. For each subsequent point (where $i \geq 2$), we check if the distance from the last reference point exceeds the predefined threshold $\tau$:
{
	\setlength\abovedisplayskip{0.2cm}
	\setlength\belowdisplayskip{0.2cm}
	\begin{equation}
		\begin{split}
             L_i - L_{last} \geq \tau 
    \end{split}	
	\end{equation}
}where $last$ denotes the index of the most recently selected reference point, and $\tau$ is the predefined threshold.

To ensure the inclusion of critical reference points, we also add the last point (if it is not already included). Subsequently, we compute the centroid $c$ of all points:
{
	\setlength\abovedisplayskip{0.2cm}
	\setlength\belowdisplayskip{0.2cm}
	\begin{equation}
		\begin{split}
             \mathbf{c} = \left( \frac{1}{k} \sum_{i=1}^k x_i, \frac{1}{k} \sum_{i=1}^k y_i \right) 
    \end{split}	
	\end{equation}
}We then add the point closest to the centroid as a reference point. Finally, ensure the sequence contains at least the starting point, the closest point to the centroid, and the end point.
The result is an ordered set of reference points $R_P$ sorted by their indices.

\textbf{Inline Sequence Attention.} 
Since our reference points $R_P$ are dynamic, especially with significant differences in the number of representative points between long and short strokes, this poses a challenge for reasonably allocating finer-grained and more matched features. Considering that the $R_p$ themselves carry positional information and can serve as a form of encoding, we designed an inline attention mechanism to feature pairing. However, before this, it is necessary to extract their sequential features.
\begin{equation}
\begin{aligned}
    &F_{p} = LSTM( Conv_{1d}(R_P) ) \\
    & F_{s} = LSTM( Conv_{1d}(S_i) )
\end{aligned}
\end{equation}where $Conv_{1d}$ represents the $1d$ convolution network, $LSTM$ indicates long-short-term memory.

Let the number of reference points within a stroke be \(D\)  and the feature channel be \(C\). We have strokes features \(F_s\) with shape \((M,C)\) and point features \(F_p\) with shape \((M \times D, C)\). As illustrated in Fig.~\ref{fig_3(a)}, to match each $R_p$ with its corresponding feature, we repeat \(F_s\) based on \(D\), resulting in \((M \times D, C)\), which represents the overall features at the stroke-level. However, inline attention requires fine-grained point-level feature allocation within each stroke. Therefore, we need to reshape both \(F_s\) and \(F_p\) into \((D, M, C)\), then linearly project them to generate the query matrix \(Q=F_p W_q\), key matrix \(K=F_s W_k\), and value matrix \(V=F_s W_v\), where \(W_q, W_k,\) and \(W_v\) are weight matrices. It is important to note that our \(D\) is dynamic. When converting the entire process into batch-level processing, certain masks are needed to assist.
{
	\setlength\abovedisplayskip{0.2cm}
	\setlength\belowdisplayskip{0.2cm}
	\begin{equation}
		\begin{split}
             R_F = Attention(Q,K,V)
            \end{split}	
	\end{equation}
}where $Attention(Q,K,V ) = V \cdot Softmax(Q^T \cdot K)$. Then, we connect the stroke sequence features through a residual connection:
\begin{equation}
\begin{aligned}
    &R_F = LayerNorm(R_F) \\
    &R_F = Relu(F_s + R_F)
\end{aligned}
\end{equation}Ultimately, each reference point and its corresponding features can be represented as the paired input $(R_P, R_F)$.

\subsection{Hierarchical Pairs Set Feature Learning}
\textbf{Hierarchical Architecture.}
Our method adopts an advanced hierarchical architecture designed to establish detailed feature associations between hierarchical reference pairs while maintaining computational efficiency. Inspired by PointNet++'s design principles, we have developed a multi-level hierarchical network that systematically abstracts increasingly larger local regions. This involves constructing hierarchical groupings of reference pairs and progressively refining these groupings to achieve more comprehensive feature representations.

The hierarchical structure consists of multiple levels of abstraction. At each level, a set of reference pairs is processed to yield a reduced yet more informative set of reference pairs. Central to our approach are four key components: Downsample, Cross-Ellipse Query, Feature Extraction, and Max Pool similar to PointNet++'s Set Abstraction module.

Initially, the Downsample Layer employs the Farthest Point Sampling (FPS) algorithm to select a subset of reference points as new centers that define local region coordinates. Following this is the Cross-Ellipse Query, which identifies "neighboring" points around each centroid to construct localized region sets as illustrated in Fig.~\ref{fig_3(b)}. The Feature Extraction phase utilizes a small PointNet-like network that interacts with these local region features capturing essential characteristics and context. Finally, Max Pool aggregates features from these local regions producing summarized representations for each set of reference pairs.

At each hierarchical level, our process consumes an \(N \times (d + C)\) matrix where \(N\) represents the number of reference points; $d$, two-dimensional coordinates; and $C$, dimensionality of reference features respectively. The output is an \(N^\prime \times (d + C^\prime)\) matrix featuring updated coordinates (d) and dimensionality \(C^\prime\) for newly generated points \(N^\prime\). This transformation allows progressive summarization of local context thereby enhancing overall performance and robustness within our network.


\textbf{ Pairs Feature Propagation for Set Segmentation.} 
In the Set Abstraction layer, we perform downsampling on the original reference pairs. To acquire semantic information for each reference pair, it is necessary to propagate features from these downsampled points back to their corresponding original reference points through interpolation.

Similar to PointNet++, we employ a hierarchical propagation strategy that utilizes distance-based interpolation and cross-level skip connections. This approach propagates features from an output pair set at abstraction level \( N_l \times (d + C) \) to an input pairs set \( N_{l-1} \), where \( N_{l-1} \) and \( N_l \) represent the sizes of the input and output pairs sets at abstraction level \( l \), respectively.

Feature propagation is achieved by interpolating feature values $f$ of $N_l$ points at coordinates corresponding to $N_{l-1}$ points. The interpolation formula is as follows:
{
	\setlength\abovedisplayskip{0.2cm}
	\setlength\belowdisplayskip{0.2cm}
	\begin{equation}
		\begin{split}
            f^{(i)}(x) = \frac{\sum_{i=1}^{k} \omega_i(x)f^{(j)}_{i} }{\sum_{i=1}^{k} \omega_i(x)} \quad \thinspace j = 1,\cdots,C
		\end{split}
	\end{equation}
}$\text{where}\thinspace \thinspace \omega_i(x) = \frac{1}{d(x,x_i)^p}$ is the inverse distance weight, $d(x,x_i)^p$ denotes $p$-th power of distance between centroid point $(x)$ and point $(x_i)$. 

After interpolating these features, they are concatenated with skip connection features retained from set abstraction level at coordinates corresponding to $ N_{l-1} $ points. These concatenated features are then passed through Multi-Layer Perceptron (MLP) for further extraction. This process repeats until all required information has been propagated back into its respective original references ensuring adequate preservation during both down-sampling and up-sampling stages.

\begin{figure}[t]
\centering
\subfigure[Before adding Aux-Branch]{
    \label{fig4(a)}    
    \includegraphics[width=3.85cm,height=4.56cm]{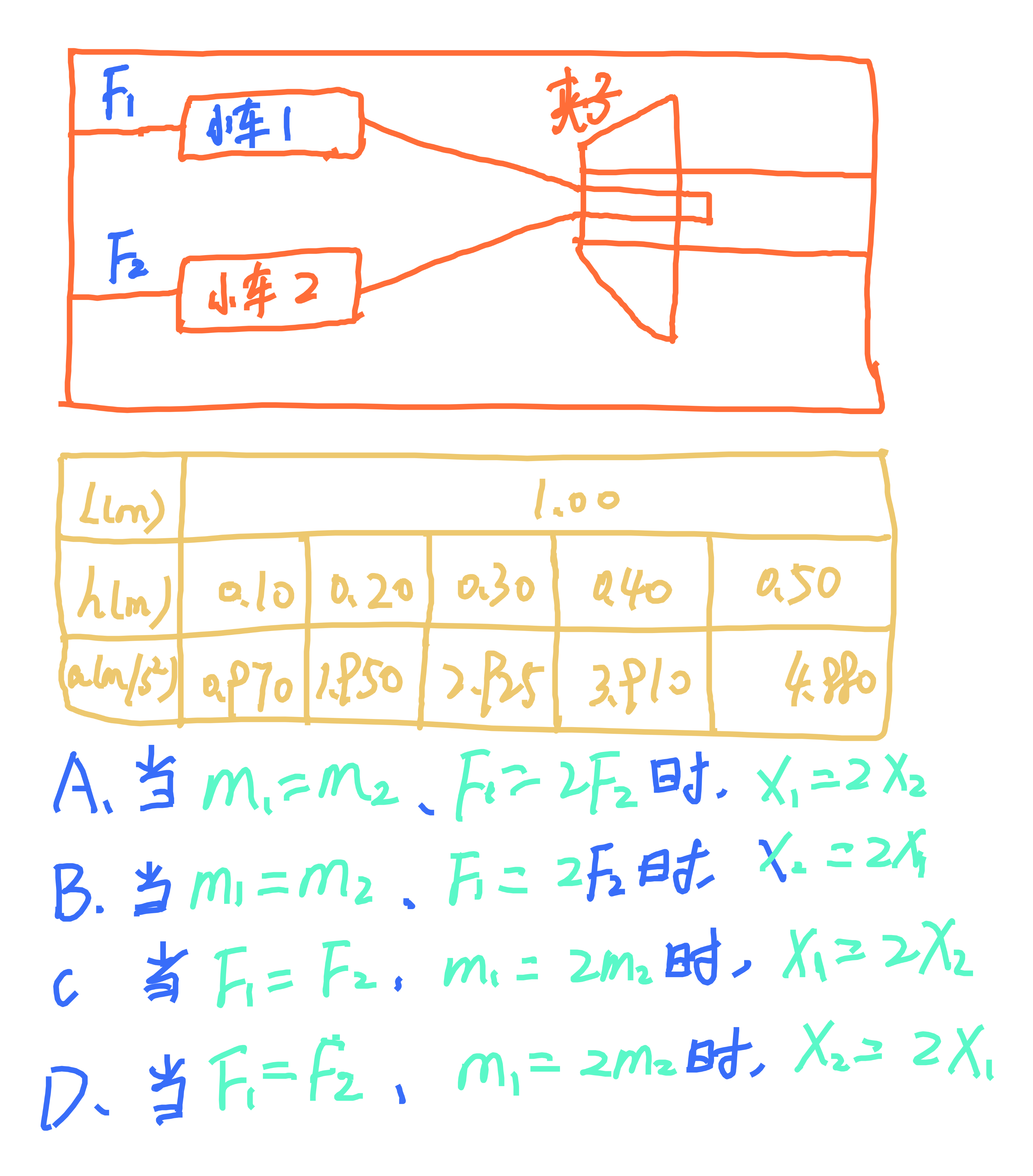}
}
\subfigure[Prediction of Aux-Branch]{
    \label{fig4(b)}  
    \includegraphics[width=3.85cm,height=4.56cm]{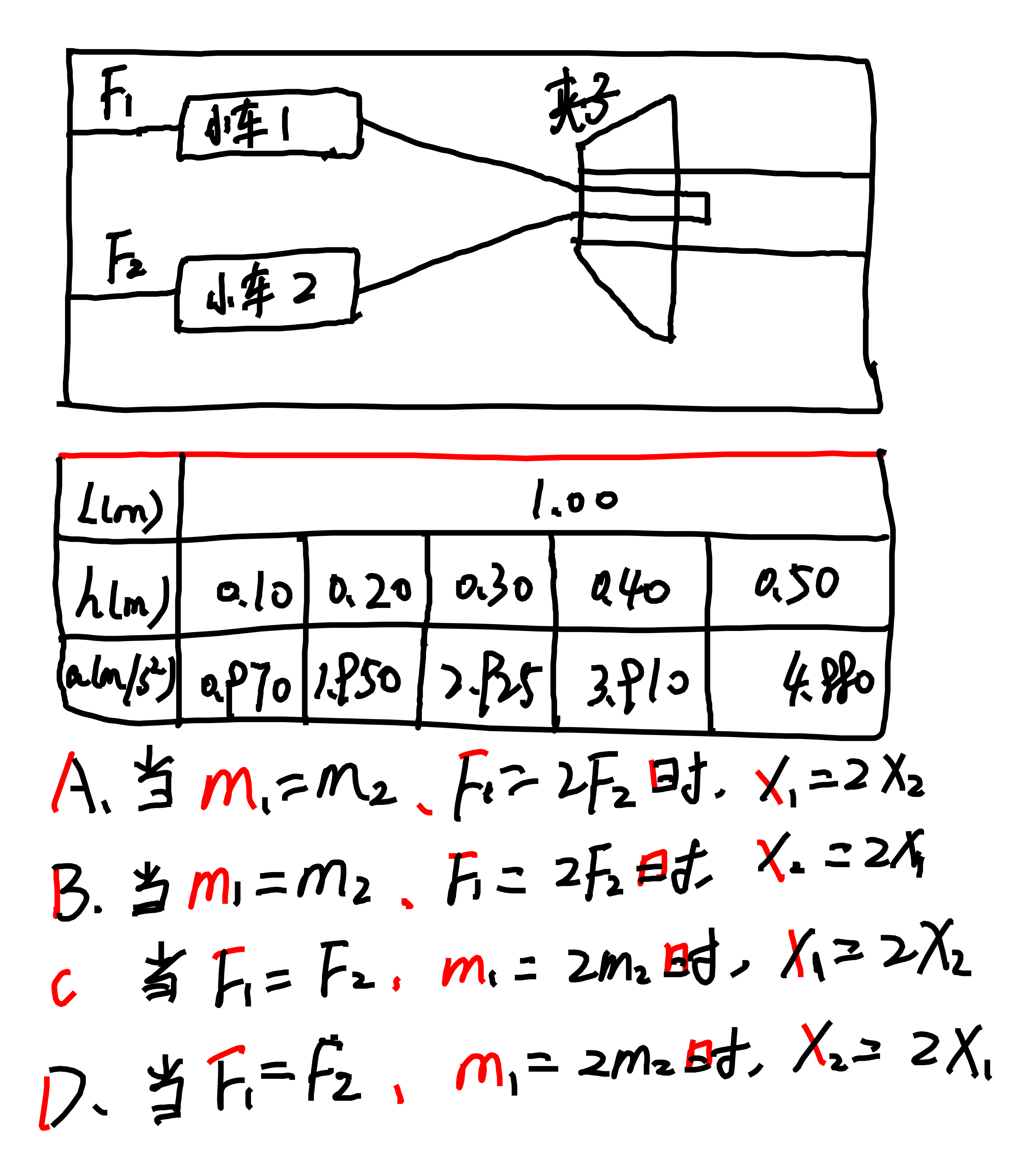}
}
\subfigure[After adding Aux-Branch]{
    \label{fig4(c)}  
    \includegraphics[width=3.85cm,height=4.56cm]{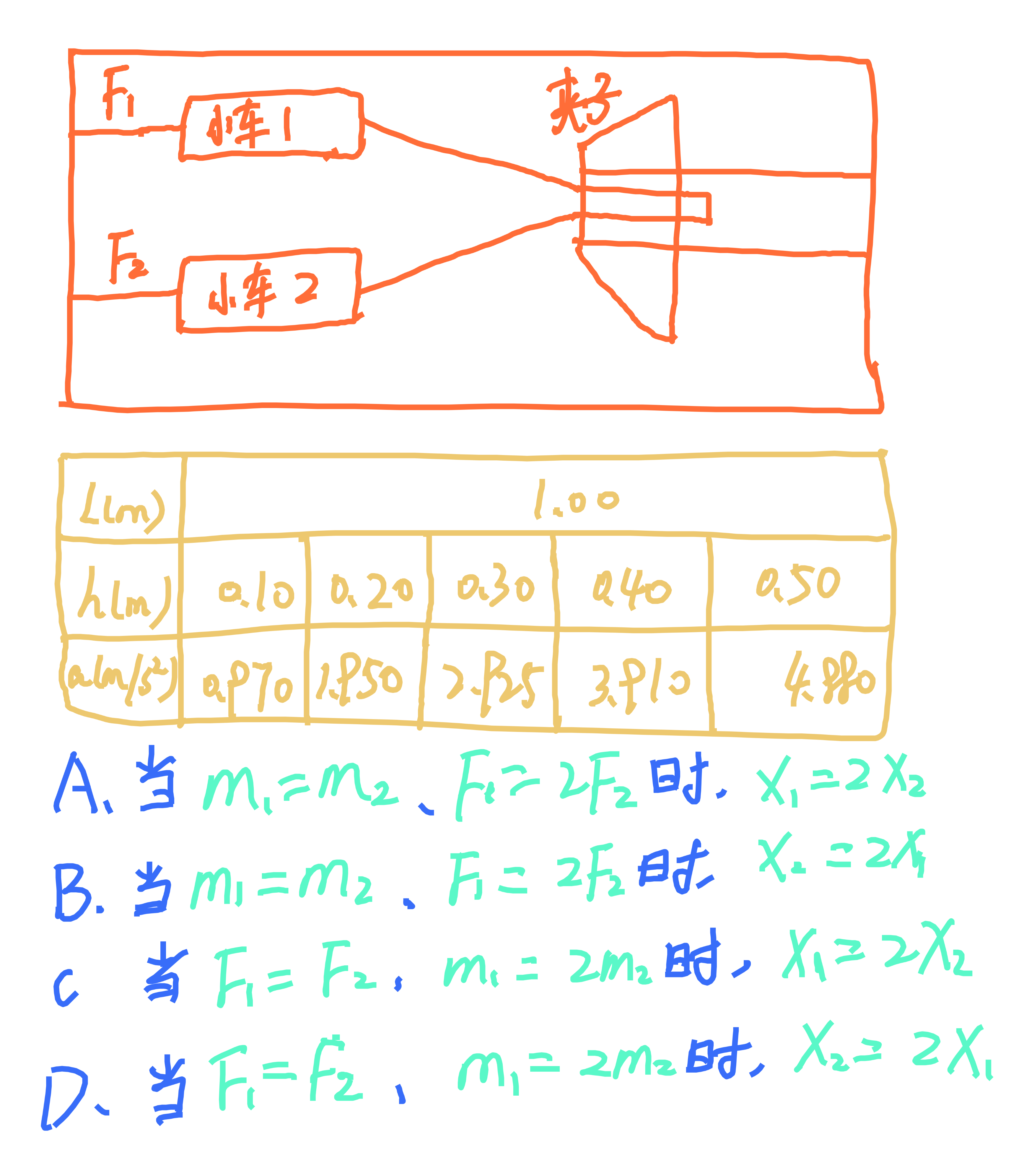}
}
\vspace{-0.8em}
\caption{ Use auxiliary branch to assist the main branch. }
\label{fig4}
\vspace{-1.2em}
\end{figure}

\subsection{Reference Pairs to Stroke Output}
Each reference point's semantic feature is obtained. However, it is still necessary to regress these features at the stroke level. Given that there is a clear mapping relationship between strokes and their corresponding pairs during the construction of dynamic reference points, this relationship can be utilized to group and construct new pair sets. These new pair sets then undergo feature extraction and pooling processes to obtain the semantic features of each stroke.

Two tasks will be performed using these final stroke features. The first task involves classifying each stroke to determine its semantic category. The second task draws inspiration from Phan et al.'s work \cite{van2016combination}, which suggests that closer proximity of two strokes within a document implies greater shared context. Conversely, interactions among more distantly separated strokes may introduce detrimental effects, such as semantic bias caused by groups constructed via CEQ. To address this issue, supervision will be applied to detect any semantic shifts between adjacent strokes.

As illustrated in Fig.~\ref{fig4}, predictions from the Aux-Branch are shown in Fig.~\ref{fig4(b)},  where red lines indicate semantic changes between consecutive strokes. By comparing Fig.~\ref{fig4(a)}  with Fig.~\ref{fig4(c)}, it becomes evident that incorporating an Aux-Branch effectively mitigates unreasonable semantic shifts between adjacent strokes.

\section{Experiments}
In this section, we introduce four publicly available online handwritten document datasets (Section~\ref{Dataset}), describe implementation details of our framework (Section~\ref{Implementation}), and present universal evaluation metrics for stroke classification (Section~\ref{Metrie}). We compare the performance of our StrokeNet with previous methods across four benchmarks for online handwritten document stroke classification (Section \ref{Compare}), specifically assessing both binary and six-class classification. To validate our method's effectiveness, we conduct ablation experiments exploring the impacts of different approaches, focusing on module design (Section \ref{Ablation}) and reference point selection (Section \ref{Effect}).

\subsection{Datasets} \label{Dataset}
\textbf{ CASIA-onDo \cite{yang2021casia}} is the largest online handwritten document analysis dataset. It consists of 2012 online documents, with strokes divided into 6 categories: Text, Formula, Diagram, Table, Figure, and List. Use CASIA-onDo to evaluate two tasks: text and multi category classification.

\textbf{IAMonDo \cite{indermuhle2010iamondo}} is a dataset used for online handwritten document layout analysis. It consists of approximately 1000 online documents, with strokes divided into 5 categories: Graphics, Text, Table, List, and Math. Perform two task evaluations using IAMonDo, one for text and the other for multi category classification.

\textbf{FC \cite{awal2011first}} is a dataset of online handwritten flow chart. It consists of 419 online process chat records. The strokes in FC are divided into 7 categories: Text, Arrow, Data, Decision, Process, Terminator, and Connection. Due to the lack of validation set in FC, we followed the approach proposed by Ye et al. \cite{ye2019contextual}.

\textbf{FA \cite{bresler2014recognition}} is a handwritten finite state automaton dataset that contains 300 charts generated by 25 authors. The strokes in FA are divided into Label, Arrow, State, and Final State.

\subsection{Implementation Details} \label{Implementation}
Our framework integrates three key components to capture fine-grained stroke dynamics, spatial hierarchies, and semantic transitions. We first encode each stroke as a sequence of reference point pairs to model detailed intra-stroke interactions. To ensure coverage of geometric variability, points are dynamically selected along the stroke based on cumulative length, with a distance threshold of 0.075, typically resulting in 3–10 points per stroke. All coordinates are normalized to the $[-1,1]$ range to eliminate scale sensitivity. These sequences are then processed using the Inline Sequence Attention (ISA) module, which models positional dependencies and enhances intra-stroke representation. To capture broader spatial context, we adopt a hierarchical grouping strategy inspired by point cloud encoders. The network progressively samples point sets of sizes $\{1024, 512, 256, 128\}$, each associated with an increasing query radius $\{0.05, 0.1, 0.2, 0.4\}$. At each level, a Cross-Ellipse Query (CEQ) operator aggregates neighboring points using elliptical search regions defined by axis ratios $\{1\!:\!5, 2\!:\!2, 5\!:\!1\}$, allowing the model to adapt to stroke structures with varying aspect ratios and orientations. Finally, to enhance the model’s ability to understand transitions between strokes, we introduce a dual-branch architecture. An auxiliary classification branch shares the encoder and predicts binary labels for adjacent stroke pairs: 0 if they belong to the same category and 1 otherwise. For $N$ strokes, $N-1$ such transition labels are generated. A weighted cross-entropy loss with a $\{1\!:\!10\}$ ratio is applied to mitigate class imbalance and promote semantic consistency across stroke boundaries. All models are trained using the Adam optimizer with an initial learning rate of 0.001. A cosine annealing schedule is employed to gradually decay the learning rate over time, facilitating stable convergence. Training is conducted on four NVIDIA RTX 4090 GPUs.

\subsection{Evaluation Metrics} \label{Metrie}
We evaluate our model on text/non-text classification and multi-class classification tasks. For text/non-text classification, accuracy as an evaluation metric is defined as follows:
{
	\setlength\abovedisplayskip{0.2cm}
	\setlength\belowdisplayskip{0.2cm}
	\begin{gather}
        \text{accuracy} =  \frac{\sum_{u=1}^{U} \sum_{m=1}^{M_u} \delta(\hat{y}_{um} = y_{um})}{ \sum_{u=1}^{U}M_u} 
	\end{gather}
}where $U$ is the number of documents and $M_u$ is the stroke count in the $u$-th document. $\hat{y}_{um}$ and  ${y}_{um}$ represent the predicted and ground truth labels, respectively. For multi-class classification, the accuracy of each class $t$ is defined as:
{
	\setlength\abovedisplayskip{0.2cm}
	\setlength\belowdisplayskip{0.2cm}
	\begin{gather}
        \text{accuracy}[t] =  \frac{\sum_{u=1}^{U} \sum_{m=1}^{M_u} \delta({y}_{um} =t) \delta(\hat{y}_{um} = y_{um})}{ \sum_{u=1}^{U}\sum_{m=1}^{M_u}\delta({y}_{um} =t)} 
	\end{gather}
}

\subsection{Comparing with State-of-the-art Methods} \label{Compare}
\textbf{Text/Nontext Classification.} 
Tables~\ref{Table:1} and~\ref{Table:2} present a comprehensive comparison of our text/nontext classification model with previous top-performing methods on the CASIA-onDo and IAMonDo datasets, respectively.
On the CASIA-onDo dataset, our model achieves an accuracy of 97.89\%, outperforming all existing single models based on graph neural networks, while avoiding reliance on manually designed features.
On the IAMonDo dataset, our model yields a 0.13\% improvement in precision compared to prior methods. Although the margin is relatively small, it indicates the model’s consistent advantage across different datasets.
In particular, our method performs better than BiLSTM and CRF-based approaches, which tend to underutilize spatial and contextual stroke patterns.
These results suggest that our model captures stroke-level information more effectively, contributing to improved classification performance without increasing model complexity.
We attribute this improvement to the integration of both local stroke attention and hierarchical context modeling, which enhances the model's ability to generalize across handwriting styles.

\begin{table}[H]
	\renewcommand{\arraystretch}{1.1}
	\vspace{-1.0em}
    \setlength{\tabcolsep}{1.8mm}{
		\begin{center}
			\caption{Text/Nontext stroke classification on the CASIA-onDo.}
			\begin{tabular}{ccc}
				\hline
				\textbf{Method} & \textbf{Description} & \textbf{Accuracy} \\
				\hline 
				EGAT \cite{ye2020contextual} &  EGAT with hand-crafted feature. & 96.12 \\
                EGAT(Auto) \cite{ye2020contextual} & EGAT with BiLSTM. & 96.72 \\
                LR-Graph \cite{yao2024long} & Long-range graph u-nets. & 97.68\\
			\textbf{StrokeNet(Ours)}  & Reference pair interaction networks.   & \textbf{97.89} \\
				\hline 
			\end{tabular}
			\label{Table:1}
	\end{center}}
	\vspace{-3.5em}
\end{table}
\begin{table}[H]
	\renewcommand{\arraystretch}{1.1}
	\setlength{\tabcolsep}{1.8mm}{
		\begin{center}
			\caption{Text/Nontext stroke classification on the IAMonDo.}
			\begin{tabular}{ccc}
				\hline
				\textbf{Method} & \textbf{Description} & \textbf{Accuracy} \\
				\hline 
                Indermühle et al. \cite{indermuhle2012mode} &  BLSTM networks. & 97.01 \\
                Delaye et al. \cite{delaye2014contextual} &   CRF with multiple contexts & 97.21 \\
                GCN \cite{ye2019contextual} &  Graph convolutional networks. & 97.21 \\
                GAT \cite{ye2019contextual} &  Graph attention networks. & 97.87 \\
                EGAT \cite{ye2020contextual} & EGAT with hand-crafted feature. & 98.65 \\
                LR-Graph \cite{yao2024long} & Long-range graph u-nets. & 98.78\\
			\textbf{StrokeNet(Ours)} & Reference pair interaction networks.   & \textbf{98.91} \\
				\hline 
			\end{tabular}
			\label{Table:2}
	\end{center}}
\end{table}
\vspace{-2.8em}

\textbf{Multi-Class Classification.} Tables~\ref{Table:3}, ~\ref{Table:4}, ~\ref{Table:5}, and ~\ref{Table:6} compare the multi-class classification results of our model on the CASIA-onDo, IAMonDo, FA, and FC datasets. Our model achieved state-of-the-art (SOTA) performances across all datasets. 
Notably, our model achieved accuracies of 95.54$\%$ and 98.45$\%$ on the complex layout data of the CASIA-onDo and IAMonDo datasets, respectively.
For the CASIA-onDo dataset, although our performance on embedded formulas and flowcharts was comparable to other methods, we saw significant improvements in other classification tasks such as tables and diagrams. This indicates that our model effectively handles diverse stroke structures within complex layouts. 
In the IAMonDo dataset, our inclusion of a Cross-Ellipse Query design facilitated high accuracy for categories with extreme aspect ratios, resulting in an overall accuracy improvement of 0.34$\%$. This demonstrates our model's ability to adapt to varying spatial dimensions within handwritten documents. 
On the relatively simpler FA and FC datasets, our model achieved accuracies of 98.98$\%$ and 99.59$\%$, respectively, demonstrating stronger contextual understanding of simple layouts compared to GCN and Transformer models. 
Specifically, on the FC dataset, the connection category achieved 100$\%$ accuracy. These results indicate that our model effectively captures relational information between strokes using reference features and successfully handles various layout characteristics of handwritten documents.

\vspace{-0.85em}
\begin{table}[H]
	\renewcommand{\arraystretch}{1.0}
	\setlength{\tabcolsep}{1.9mm}{
		\begin{center}
			\caption{Multi-Class stroke classification on the CASIA-onDo.}
			\begin{tabular}{cccccccc}
				\hline
				\textbf{Method} & \textbf{Text}& \textbf{Formula}& \textbf{Diagram}& \textbf{Table} & \textbf{Figure} & \textbf{List}
				& \textbf{Accuracy} \\
				\hline 
				EGAT \cite{ye2020contextual} & 89.59 & 76.94 & 98.17 & 94.50 & 91.26 & 76.13  & 88.79\\
				EGAT(Auto) \cite{ye2020contextual} & 93.04 & \textbf{89.43} & 96.07 & 94.35 & 91.17 & 68.55 & 89.87\\
				LR-Graph \cite{yao2024long} & 94.26 & 86.06 & 98.14  & 96.71 & 91.41 & 90.05  & 93.56\\
				T-OHS \cite{liu2024transformer}& 95.07 & 88.08 & \textbf{98.64} & 96.02 & 93.00 & 90.94 & 93.81\\
				\textbf{StrokeNet(Ours)} & \textbf{96.34} & 89.20 & 97.11 & \textbf{97.31} & \textbf{95.38} & \textbf{94.78} & \textbf{95.54}\\
				\hline 
			\end{tabular}
			\label{Table:3}
	\end{center}}
\end{table}
\vspace{-4.4em}
\begin{table}[H]
	\renewcommand{\arraystretch}{1.0}
	\setlength{\tabcolsep}{1.9mm}{
		\begin{center}
			\caption{Multi-Class stroke classification on the IAMonDo.}
			\begin{tabular}{ccccccc}
				\hline
				\textbf{Method} & \textbf{Graphics} & \textbf{Text}& \textbf{Table}& \textbf{List}& \textbf{Math} & \textbf{Accuracy} \\
				\hline 
				GCN \cite{ye2019contextual}& 93.09 &  97.97 &  70.31 & 52.73 &  74.48 &  91.11\\
				GAT \cite{ye2019contextual}&  95.12 & 98.19 & 77.59 & 69.66 & 82.22 &  93.51 \\
				EGAT \cite{ye2020contextual} & 97.11 & 98.35 & 89.70  &  76.15 & 88.43 & 95.81 \\
                LR-Graph \cite{yao2024long} & 98.00 & 98.83 & 97.12  & \textbf{98.80} &  85.65 & 97.89 \\
				T-OHS \cite{liu2024transformer} & 97.88 & 99.31 & 98.11 & 87.86 & \textbf{94.89} & 98.11 \\
				\textbf{StrokeNet(Ours)} & \textbf{98.86} & \textbf{99.31} & \textbf{98.29} & 90.29 & 93.59 & \textbf{98.45}\\
				\hline 
			\end{tabular}
			\label{Table:4}
	\end{center}}
\end{table}
\vspace{-4.05em}
\begin{table}[H]
	\renewcommand{\arraystretch}{1.0}
	\setlength{\tabcolsep}{0.9mm}{
		\begin{center}
			\caption{Multi-Class stroke classification on the FC.}
            \vspace{-1.6em}
			\begin{tabular}{ccccccccc}
				\hline
				\textbf{Method} & \textbf{Text}& \textbf{Arrow}& \textbf{Data}& \textbf{Decision} & \textbf{Process} & \textbf{Terminator} &\textbf{Connection}
				& \textbf{Accuracy} \\
				\hline 
				GCN \cite{ye2019contextual}& 97.74 & 89.44 & 88.67 & 92.92 & 82.30 & 85.96  & 85.19 & 93.99\\
				GAT \cite{ye2019contextual}& 98.01 & 88.21 & 88.57 & 92.27 & 82.39 & 87.09 & 87.56 & 94.00\\
                    EGAT \cite{ye2020contextual}& 98.84 & 96.39 & 94.40 & 95.19 & 92.94 & 93.23 & 92.89 & 97.36\\
				LR-Graph \cite{yao2024long}& 99.39 & 96.77 & 94.90 & 95.82 & 96.00 & 91.69  & 86.57 & 97.94\\
				T-OHS \cite{liu2024transformer}& \textbf{99.66} & 98.50 & \textbf{96.20} & \textbf{98.51} & \textbf{98.17} & 97.07 & 99.11 & 98.87\\
				\textbf{StrokeNet(Ours)} & 99.60 & \textbf{98.75} & 95.87 & 98.27 & 97.86 & \textbf{97.08} & \textbf{100.00} & \textbf{98.98}\\
				\hline 
			\end{tabular}
			\label{Table:5}
	\end{center}}
\end{table}
\vspace{-4.75em}
\begin{table}[H]
	\renewcommand{\arraystretch}{1.0}
	\setlength{\tabcolsep}{2.1mm}{
		\begin{center}
			\caption{Multi-Class stroke classification on the FA.}
            \vspace{-0.45em}
			\begin{tabular}{cccccc} 
				\hline
				\textbf{Method} & \textbf{Label} & \textbf{Arrow}& \textbf{State}& \textbf{Final State}& \textbf{Accuracy} \\
				\hline 
				GCN \cite{ye2019contextual}& 94.04 & 91.75 & 84.18 & 87.81 & 92.13\\
				GAT \cite{ye2019contextual}& 99.01 & 97.89 & 92.30 & 94.81 & 97.87 \\
				EGAT \cite{ye2020contextual}& 99.49 & 99.08 & 97.18 & 97.42 & 99.05\\
                LR-Graph \cite{yao2024long}& 99.76 & \textbf{99.93} & 96.52  & \textbf{98.08} & 99.49 \\
				\textbf{StrokeNet(Ours)} & \textbf{99.90} & 99.67 & \textbf{99.30} & 96.92 & \textbf{99.59}\\
				\hline 
			\end{tabular}
			\label{Table:6}
	\end{center}}
\end{table}
\vspace{-4.0em}

\subsection{Ablation Studies} \label{Ablation}
We performed ablation experiments to identify key factors influencing our model, focusing on four aspects: Inline Sequence Attention (ISA), Cross-Ellipse Query (CEQ), Reference Pairs to Stroke (RPTS), and Auxiliary Branch (Aux-Branch). For ISA, we assigned identical stroke features to each point. For CEQ, we used only individual Ball Query. For RPTS, we concatenated only the point features for stroke regression. Lastly, for Aux-Branch, we evaluated performance with/without this auxiliary branch.

We conducted comparative multi-class classification experiments on the CASIA-onDo, and the specific experimental results are presented in Table~\ref{Table:7}. The presence of effective ISA construction has a significant impact on the interaction between strokes. This is particularly evident in categories with many long strokes, where performance noticeably declines. Conversely, CEQ has a significant effect on categories with highly asymmetric dimensions, such as formulas and text lines. Compared to purely stacked features, RPTS extracts global information within strokes during regression, which helps mitigate interference from certain outlier points and leads to overall improvement. The Aux-Branch has a relatively small impact on overall accuracy compared to other modules, as semantic anomalies mainly affect the perceptual experience.
\vspace{-1.45em}
\begin{table}[H]
	\renewcommand{\arraystretch}{1.0}
	\setlength{\tabcolsep}{1.6mm}{
		\begin{center}
			\caption{The ablation experiment on the CASIA-onDo.}
            \vspace{-1.55em}
			\begin{tabular}{cccccccc} 
				\hline
				\textbf{Method} & \textbf{Text}& \textbf{Formula}& \textbf{Diagram}& \textbf{Table} & \textbf{Figure} & \textbf{List}
				& \textbf{Accuracy} \\
				\hline 
				StrokeNet(ISA) & 95.85 & 86.36 & 96.13 & 96.34 & 92.65 & 92.78  & 94.25\\
                StrokeNet(CEQ) & 95.31 & 86.34 & 95.22 & 96.60 & 92.03 & 92.55 & 93.95\\
				StrokeNet(RPTS) & 96.12 & 88.18 & 96.64 & \textbf{97.61} & 94.89 & 94.20 & 95.25\\
				StrokeNet(Aux-Branch) & 96.34 & \textbf{90.27} & 96.97  & 97.08 & 93.31 & 93.92  & 95.34\\
				\textbf{StrokeNet(Ours)} & \textbf{96.34} & 89.20 & \textbf{97.11} & 97.31 & \textbf{95.38} & \textbf{94.78} & \textbf{95.54}\\
				\hline 
			\end{tabular}
			\label{Table:7}
	\end{center}}
    \vspace{-2.8em}
\end{table}

\subsection{Effect of Different Numbers of Reference Points}  \label{Effect}
To further validate the impact of reference point selection on model performance, we configured the model with different reference point settings: StrokeNet (One), StrokeNet (Three), StrokeNet (Five), StrokeNet (Seven), and StrokeNet (Total), selecting 1, 3, 5, and 7 reference points per stroke respectively. We conducted multi-class classification experiments on the CASIA-onDo, with results shown in Table~\ref{Table:8}.

\begin{figure}[t]
	\begin{center}
		\subfigure[StrokeNet(One)]{
			\label{fig_5(a)}
			{\includegraphics[width=5.7cm,height=3.45cm]{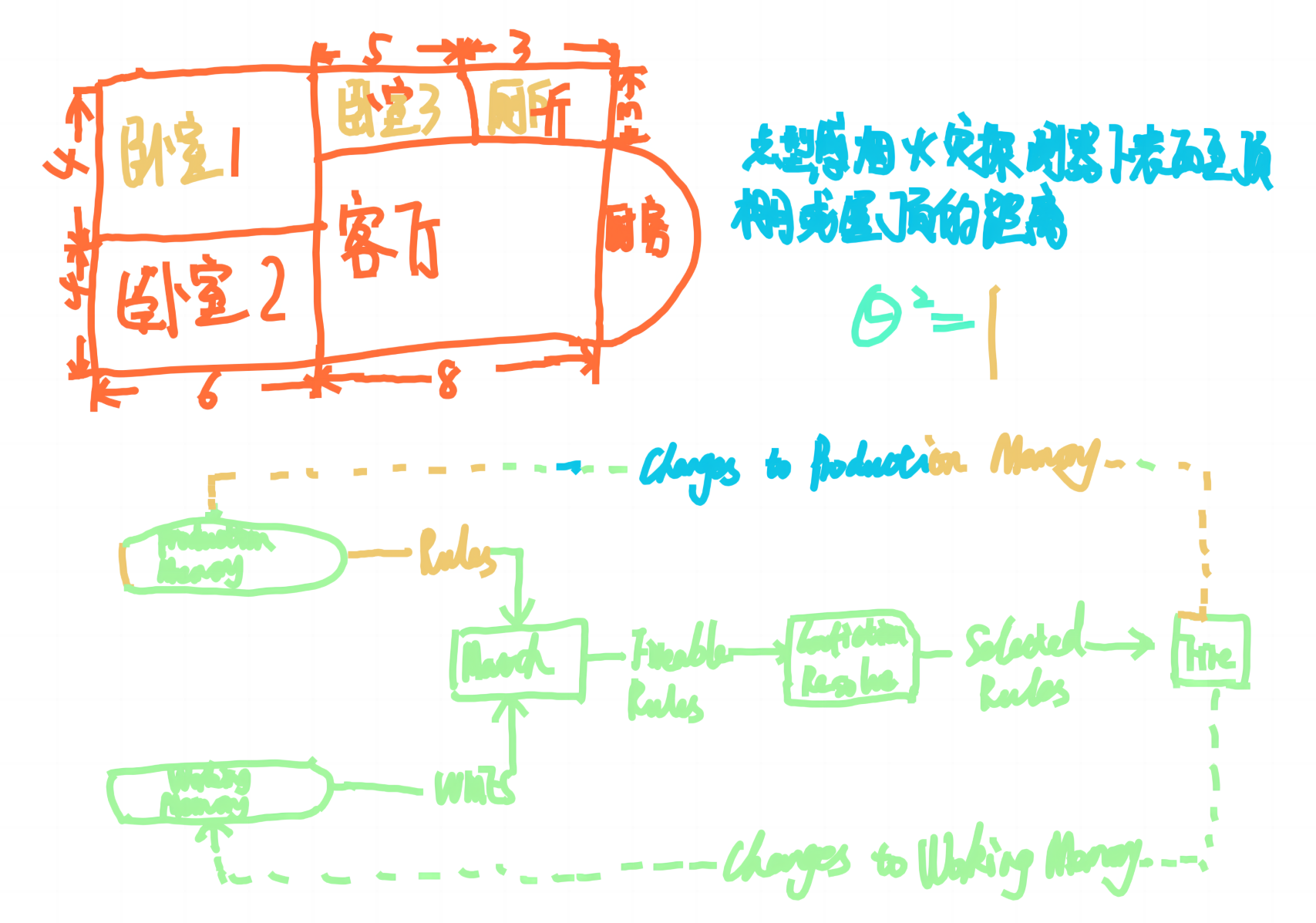}}    
		}
		\subfigure[StrokeNet(Total)]{
			\label{fig_5(b)}
			{\includegraphics[width=5.7cm,height=3.45cm]{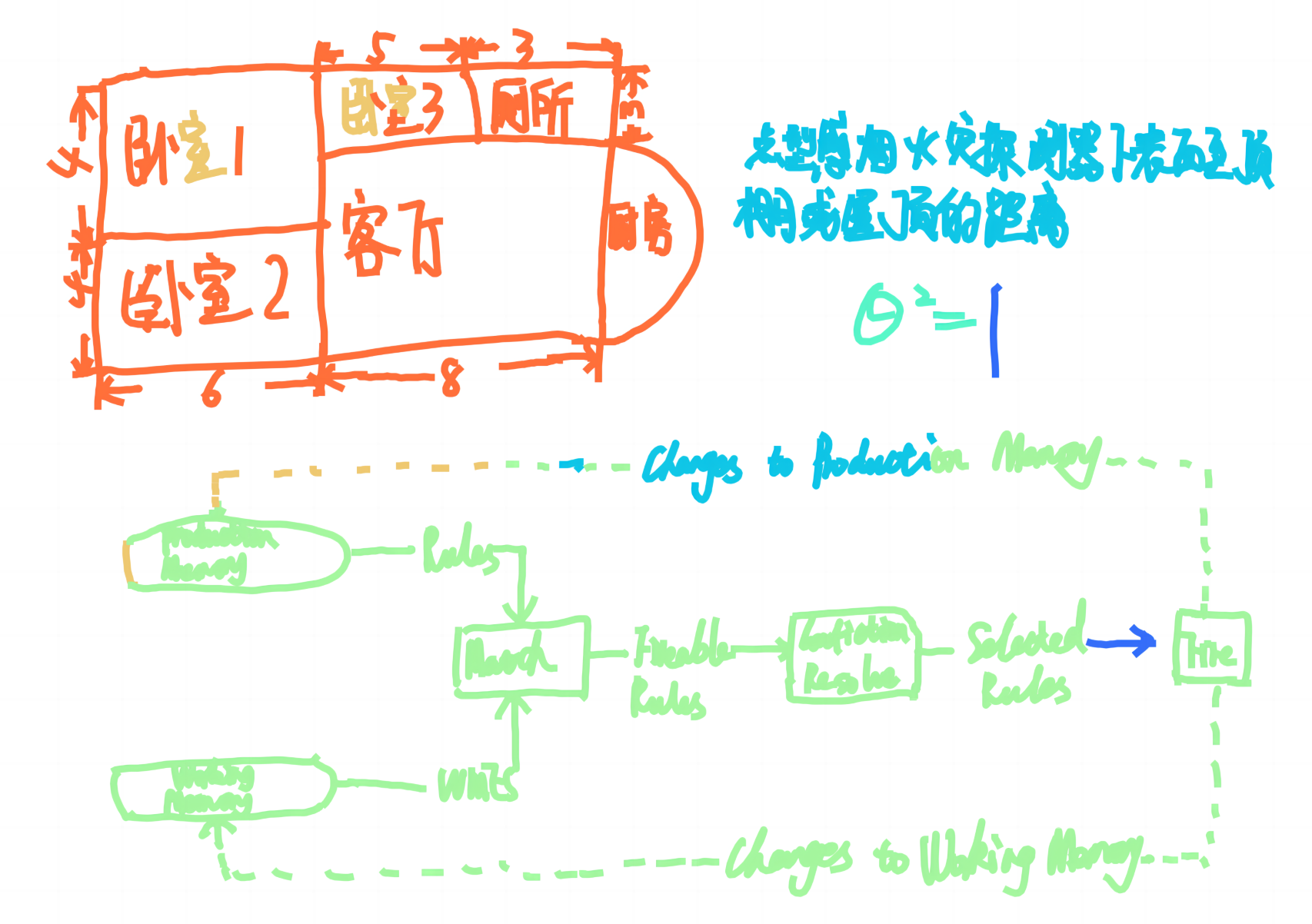}}   
		}\vspace{-0.8em}
        \subfigure[StrokeNet(Ours)]{
			\label{fig_5(c)}{\includegraphics[width=5.7cm,height=3.45cm]{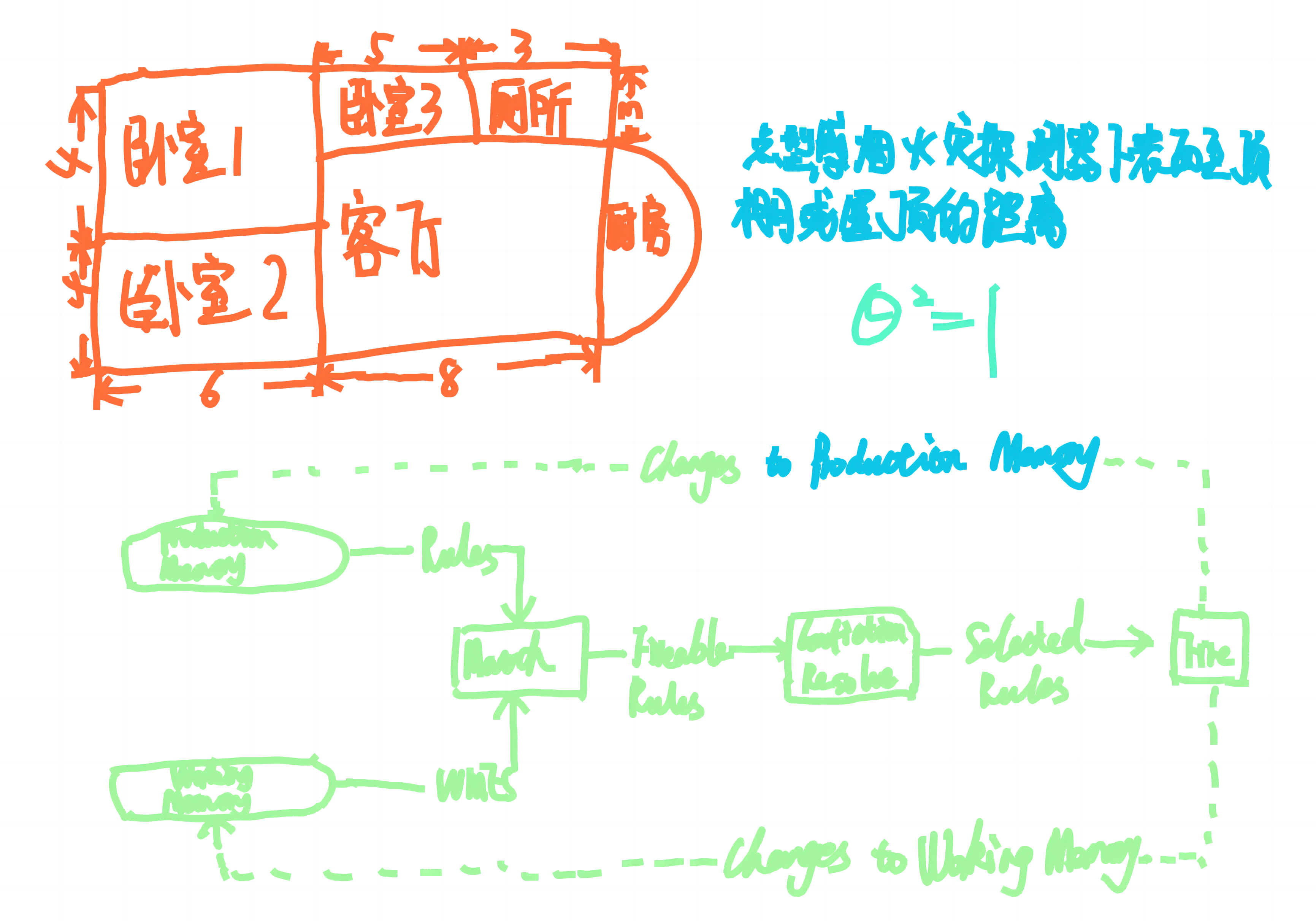}}   
		}
        \subfigure[Ground Truth]{
			\label{fig_5(d)}
			{\includegraphics[width=5.7cm,height=3.45cm]{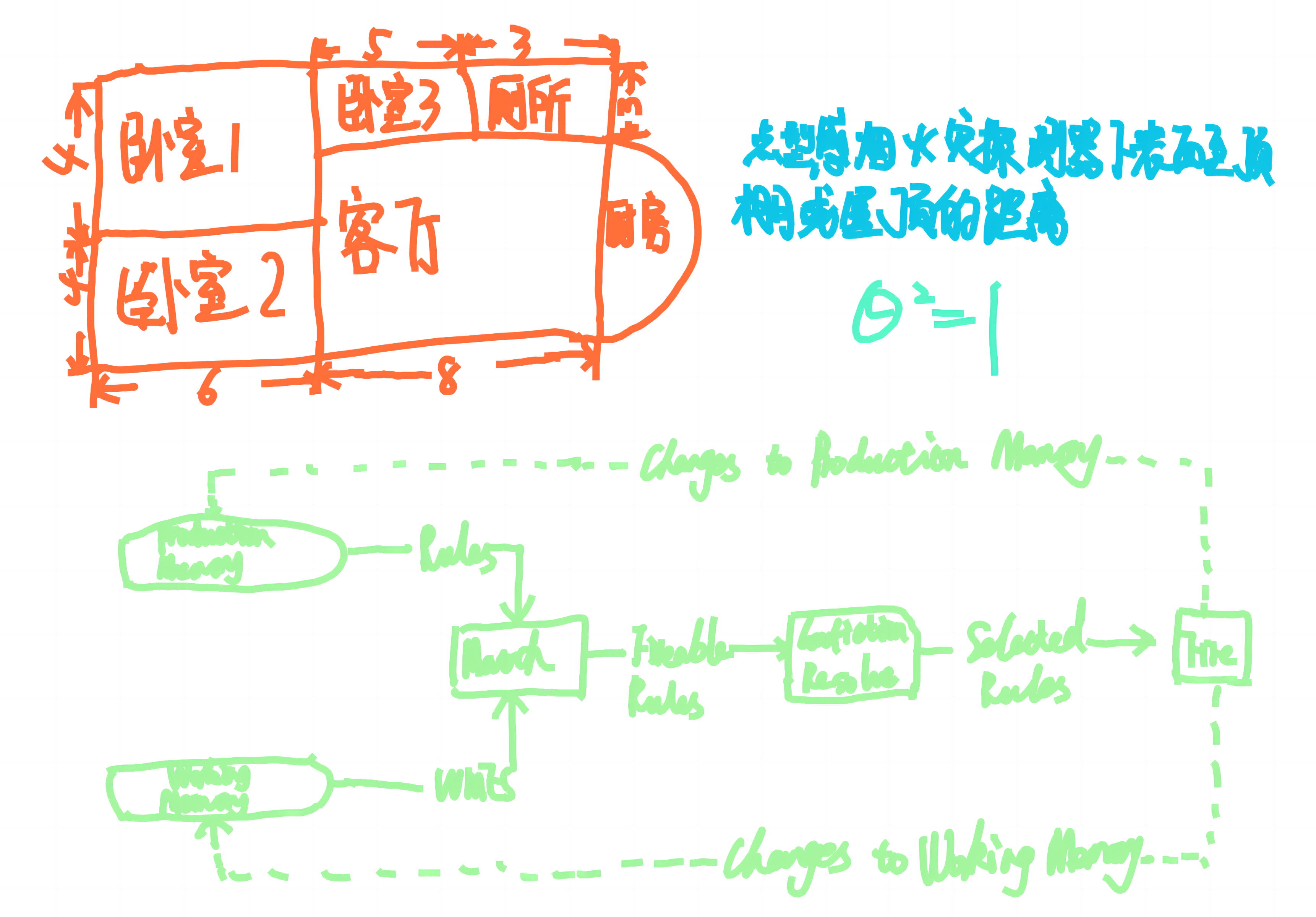}}   
		}
	\end{center}
    \vspace{-1.75em}
	\caption{The recognition results of different numbers of reference points.} \label{fig5}
    \vspace{-2.0em}
\end{figure}

The results indicate that the number of reference points significantly affects the model's performance. StrokeNet (One) has the lowest accuracy at 93.00$\%$, as shown in Fig.~\ref{fig_5(a)}, indicating that a single reference point cannot capture stroke structural features. As the number of reference points increases, both StrokeNet (Three) and StrokeNet (Five) show significant improvements in long-stroke categories such as Table and Diagram, indicating that they can better represent stroke details. However, when too many reference points are used, StrokeNet (Seven) and StrokeNet (Total) lead to decreased performance in short-stroke categories such as Text and Formula, as shown in Fig.~\ref{fig_5(b)}. This is due to information redundancy causing attention bias in the model. As illustrated in Fig.~\ref{fig_5(c)}, our dynamic selection strategy achieves a better balance between capturing sufficient stroke information and avoiding redundancy. It performs better than other configurations in most categories and significantly improves overall accuracy.

\vspace{-1.35em}
\begin{table}[H]
	\renewcommand{\arraystretch}{1.0}   
	\setlength{\tabcolsep}{1.9mm}{
		\begin{center}
			\caption{The effect of different numbers of reference points on the CASIA-onDo.}
            \vspace{-0.35em}
			\begin{tabular}{cccccccc} 
				\hline
				\textbf{Method} & \textbf{Text}& \textbf{Formula}& \textbf{Diagram}& \textbf{Table} & \textbf{Figure} & \textbf{List}
				& \textbf{Accuracy} \\
				\hline 
				StrokeNet(One) & 94.34 & 87.39 & 96.13 & 96.45 & 88.78 & 88.13  & 93.00\\
                StrokeNet(Three) & \textbf{96.58} & 88.85 & 96.94 & 96.05 & 90.22 & 91.09 & 94.42\\
				StrokeNet(Five) & 96.13 & 88.25 & 97.02 & 97.12 & 92.56 & 91.29  & 94.64\\
                StrokeNet(Seven) & 95.83 & 84.94 &  \textbf{97.25} & 97.25 & 91.12 & 94.66 & 94.69\\
			StrokeNet(Total) & 95.13 & 83.12 & 94.32 & 95.84 & 90.71 & 92.37 & 93.12\\
                \textbf{StrokeNet(Ours)} & 96.34 & \textbf{89.20} & 97.11 & \textbf{97.31} & \textbf{95.38} & \textbf{94.78} & \textbf{95.54}\\
				\hline 
			\end{tabular}
			\label{Table:8}
	\end{center}}
\end{table}
\vspace{-4.5em}

\section{Conclusion and Future Work}
In this paper, we propose a novel paradigm for handling online handwritten stroke classification through reference pairs. Specifically, our method designs a dynamic adaptive mechanism to select reference points and Inline Sequence Attention to construct reference point-feature pairs. In the hierarchical network, we design Cross-Ellipse Query to adapt to writing habits and facilitate interactions between strokes. To prevent semantic mutations, we design an Auxiliary Branch for supervision. We conducted experiments on online handwritten document datasets such as CASIA-onDo, IAMonDo, FC, and FA. The results demonstrate that our method significantly outperforms previous approaches. Although these results are encouraging, our framework still faces challenges in higher-level semantic tasks such as content detection and understanding. Future work will focus on addressing these challenges and further extending our approach.

\section{Acknowledgement} 
This work is supported by the National Natural Science Foundation of China (No. 91748107), the foundation of State Key Laboratory of Public Big Data(No.PBD2023-11), the Guangdong Innovative Research Team Program (No. 2014ZT05G157).

\vspace{-1.15em}
%
%

\bibliographystyle{splncs04}
\bibliography{reference.bib}  

@article{artz2020taking,
  title={Taking notes in the digital age: Evidence from classroom random control trials},
  author={Artz, Benjamin and Johnson, Marianne and Robson, Denise and Taengnoi, Sarinda},
  journal={The Journal of Economic Education},
  volume={51},
  number={2},
  pages={103--115},
  year={2020},
  publisher={Taylor \& Francis}
}

@inproceedings{polotskyi2020spatio,
  title={Spatio-temporal clustering for grouping in online handwriting document layout analysis with GRU-RNN},
  author={Polotskyi, Serhii and Radyvonenko, Olga and Degtyarenko, Illya and Deriuga, Ivan},
  booktitle={2020 17th International Conference on Frontiers in Handwriting Recognition (ICFHR)},
  pages={276--281},
  year={2020},
  organization={IEEE}
}

@inproceedings{ye2016joint,
  title={Joint training of conditional random fields and neural networks for stroke classification in online handwritten documents},
  author={Ye, Jun-Yu and Zhang, Yan-Ming and Liu, Cheng-Lin},
  booktitle={2016 23rd International Conference on Pattern Recognition (ICPR)},
  pages={3264--3269},
  year={2016},
  organization={IEEE}
}

@inproceedings{indermuhle2010iamondo,
  title={IAMonDo-database: an online handwritten document database with non-uniform contents},
  author={Inderm{\"u}hle, Emanuel and Liwicki, Marcus and Bunke, Horst},
  booktitle={Proceedings of the 9th IAPR international workshop on document analysis systems},
  pages={97--104},
  year={2010}
}

@inproceedings{awal2011first,
  title={First experiments on a new online handwritten flowchart database},
  author={Awal, Ahmad-Montaser and Feng, Guihuan and Mouchere, Harold and Viard-Gaudin, Christian},
  booktitle={Document Recognition and Retrieval XVIII},
  volume={7874},
  pages={81--90},
  year={2011},
  organization={SPIE}
}

@inproceedings{bresler2014recognition,
  title={Recognition system for on-line sketched diagrams},
  author={Bresler, Martin and Van Phan, Truyen and Prusa, Daniel and Nakagawa, Masaki and Hlav{\'a}c, V{\'a}clav},
  booktitle={2014 14th International Conference on Frontiers in Handwriting Recognition},
  pages={563--568},
  year={2014},
  organization={IEEE}
}

@inproceedings{yang2021casia,
  title={CASIA-onDo: a new database for online handwritten document analysis},
  author={Yang, Yu-Ting and Zhang, Yan-Ming and Yun, Xiao-Long and Yin, Fei and Liu, Cheng-Lin},
  booktitle={Asian Conference on Pattern Recognition},
  pages={174--188},
  year={2021},
  organization={Springer}
}

@inproceedings{bishop2004distinguishing,
  title={Distinguishing text from graphics in on-line handwritten ink},
  author={Bishop, Christopher M and Svensen, Markus and Hinton, Geoffrey E},
  booktitle={Ninth International Workshop on Frontiers in Handwriting Recognition},
  pages={142--147},
  year={2004},
  organization={IEEE}
}

@inproceedings{zhou2007text,
  title={Text/non-text ink stroke classification in japanese handwriting based on markov random fields},
  author={Zhou, X-D and Liu, C-L},
  booktitle={Ninth International Conference on Document Analysis and Recognition (ICDAR 2007)},
  volume={1},
  pages={377--381},
  year={2007},
  organization={IEEE}
}

@article{delaye2014contextual,
  title={Contextual text/non-text stroke classification in online handwritten notes with conditional random fields},
  author={Delaye, Adrien and Liu, Cheng-Lin},
  journal={Pattern Recognition},
  volume={47},
  number={3},
  pages={959--968},
  year={2014},
  publisher={Elsevier}
}

@inproceedings{indermuhle2012mode,
  title={Mode detection in online handwritten documents using BLSTM neural networks},
  author={Inderm{\"u}hle, Emanuel and Frinken, Volkmar and Bunke, Horst},
  booktitle={2012 International Conference on Frontiers in Handwriting Recognition},
  pages={302--307},
  year={2012},
  organization={IEEE}
}

@article{van2016combination,
  title={Combination of global and local contexts for text/non-text classification in heterogeneous online handwritten documents},
  author={Van Phan, Truyen and Nakagawa, Masaki},
  journal={Pattern Recognition},
  volume={51},
  pages={112--124},
  year={2016},
  publisher={Elsevier}
}

@inproceedings{grygoriev2021hcrnn,
  title={HCRNN: a novel architecture for fast online handwritten stroke classification},
  author={Grygoriev, Andrii and Degtyarenko, Illya and Deriuga, Ivan and Polotskyi, Serhii and Melnyk, Volodymyr and Zakharchuk, Dmytro and Radyvonenko, Olga},
  booktitle={International Conference on Document Analysis and Recognition},
  pages={193--208},
  year={2021},
  organization={Springer}
}

@inproceedings{ye2019contextual,
  title={Contextual stroke classification in online handwritten documents with graph attention networks},
  author={Ye, Jun-Yu and Zhang, Yan-Ming and Yang, Qing and Liu, Cheng-Lin},
  booktitle={2019 International conference on document analysis and recognition (ICDAR)},
  pages={993--998},
  year={2019},
  organization={IEEE}
}

@article{ye2020contextual,
  title={Contextual stroke classification in online handwritten documents with edge graph attention networks},
  author={Ye, Jun-Yu and Zhang, Yan-Ming and Yang, Qing and Liu, Cheng-Lin},
  journal={SN Computer Science},
  volume={1},
  pages={1--13},
  year={2020},
  publisher={Springer}
}

@inproceedings{yao2024long,
  title={Long-Range Graph U-Nets: Node and Edge Clustering Pooling Model For Stroke Classification in Online Handwritten Documents},
  author={Yao, Muwu and She, Shuang and Li, Jinrong and Lin, Jianmin and Yang, Ming and Peng, Hongxing},
  booktitle={Asian Conference on Machine Learning},
  pages={1542--1557},
  year={2024},
  organization={PMLR}
}

@article{mustafid2023iamonsense,
  title={IAMonSense: multi-level handwriting classification using spatiotemporal information},
  author={Mustafid, Ahmad and Younas, Junaid and Lukowicz, Paul and Ahmed, Sheraz},
  journal={International Journal on Document Analysis and Recognition (IJDAR)},
  volume={26},
  number={3},
  pages={303--319},
  year={2023},
  publisher={Springer}
}

@article{zheng2023sketch,
  title={Sketch-segformer: Transformer-based segmentation for figurative and creative sketches},
  author={Zheng, Yixiao and Xie, Jiyang and Sain, Aneeshan and Song, Yi-Zhe and Ma, Zhanyu},
  journal={IEEE transactions on image processing},
  volume={32},
  pages={4595--4609},
  year={2023},
  publisher={IEEE}
}

@article{liu2024transformer,
  title={Transformer-based stroke relation encoding for online handwriting and sketches},
  author={Liu, Jing-Yu and Zhang, Yan-Ming and Yin, Fei and Liu, Cheng-Lin},
  journal={Pattern Recognition},
  volume={148},
  pages={110131},
  year={2024},
  publisher={Elsevier}
}

@inproceedings{qi2017pointnet,
  title={Pointnet: Deep learning on point sets for 3d classification and segmentation},
  author={Qi, Charles R and Su, Hao and Mo, Kaichun and Guibas, Leonidas J},
  booktitle={Proceedings of the IEEE conference on computer vision and pattern recognition},
  pages={652--660},
  year={2017}
}

@article{qi2017pointnet++,
  title={Pointnet++: Deep hierarchical feature learning on point sets in a metric space},
  author={Qi, Charles Ruizhongtai and Yi, Li and Su, Hao and Guibas, Leonidas J},
  journal={Advances in neural information processing systems},
  volume={30},
  year={2017}
}

@inproceedings{namboodiri2004robust,
  title={Robust Segmentation of Unconstrained Online Handwritten Documents.},
  author={Namboodiri, Anoop M and Jain, Anil K},
  booktitle={ICVGIP},
  pages={165--170},
  year={2004}
}

@article{simonyan2014very,
  title={Very deep convolutional networks for large-scale image recognition},
  author={Simonyan, Karen and Zisserman, Andrew},
  journal={arXiv preprint arXiv:1409.1556},
  year={2014}
}

@inproceedings{zhao2023divide,
  title={Divide and conquer: 3d point cloud instance segmentation with point-wise binarization},
  author={Zhao, Weiguang and Yan, Yuyao and Yang, Chaolong and Ye, Jianan and Yang, Xi and Huang, Kaizhu},
  booktitle={Proceedings of the IEEE/CVF International Conference on Computer Vision},
  pages={562--571},
  year={2023}
}

@inproceedings{sun2023superpoint,
  title={Superpoint transformer for 3d scene instance segmentation},
  author={Sun, Jiahao and Qing, Chunmei and Tan, Junpeng and Xu, Xiangmin},
  booktitle={Proceedings of the AAAI Conference on Artificial Intelligence},
  volume={37},
  pages={2393--2401},
  year={2023}
}

@inproceedings{jiang2020pointgroup,
  title={Pointgroup: Dual-set point grouping for 3d instance segmentation},
  author={Jiang, Li and Zhao, Hengshuang and Shi, Shaoshuai and Liu, Shu and Fu, Chi-Wing and Jia, Jiaya},
  booktitle={Proceedings of the IEEE/CVF conference on computer vision and Pattern recognition},
  pages={4867--4876},
  year={2020}
}

@inproceedings{yang2019std,
  title={Std: Sparse-to-dense 3d object detector for point cloud},
  author={Yang, Zetong and Sun, Yanan and Liu, Shu and Shen, Xiaoyong and Jia, Jiaya},
  booktitle={Proceedings of the IEEE/CVF international conference on computer vision},
  pages={1951--1960},
  year={2019}
}

@article{chen2021pointnet++,
  title={PointNet++ network architecture with individual point level and global features on centroid for ALS point cloud classification},
  author={Chen, Yang and Liu, Guanlan and Xu, Yaming and Pan, Pai and Xing, Yin},
  journal={Remote Sensing},
  volume={13},
  number={3},
  pages={472},
  year={2021},
  publisher={MDPI}
}

@article{ma2022rethinking,
  title={Rethinking network design and local geometry in point cloud: A simple residual MLP framework},
  author={Ma, Xu and Qin, Can and You, Haoxuan and Ran, Haoxi and Fu, Yun},
  journal={arXiv preprint arXiv:2202.07123},
  year={2022}
}

@article{qian2022pointnext,
  title={Pointnext: Revisiting pointnet++ with improved training and scaling strategies},
  author={Qian, Guocheng and Li, Yuchen and Peng, Houwen and Mai, Jinjie and Hammoud, Hasan and Elhoseiny, Mohamed and Ghanem, Bernard},
  journal={Advances in neural information processing systems},
  volume={35},
  pages={23192--23204},
  year={2022}
}

\end{document}